\documentclass[runningheads]{llncs}

\usepackage{eccv}

\usepackage{eccvabbrv}

\usepackage{graphicx}
\usepackage{booktabs}

\usepackage[accsupp]{axessibility}  

\usepackage{hyperref}

\usepackage{orcidlink}

\usepackage{kotex}
\usepackage{booktabs}
\usepackage{amsmath, amsfonts}
\usepackage[usenames,dvipsnames,table]{xcolor}
\usepackage{algpseudocode}
\usepackage{multirow}
\usepackage{longtable}
\usepackage{arydshln}
\usepackage{pifont}
\usepackage{comment}
\usepackage[algo2e]{algorithm2e} 
\usepackage{algorithm}
\usepackage{adjustbox}
\usepackage{caption}
\usepackage{wrapfig}
\usepackage{tikz}
\usepackage{makecell}

\newcommand{\bl}{$\;\,$}
\definecolor{test}{HTML}{00F9DE}
\definecolor{DYellow}{HTML}{CCCC00}
\definecolor{DDYellow}{HTML}{430486}
\definecolor{DOrange}{HTML}{E97132}
\definecolor{Lightblue}{HTML}{215F9A}
\definecolor{DYellow2}{HTML}{F6EA00}

\newcommand{\graycb}[1]{\cellcolor[HTML]{B8B8B8}{\textbf{#1}}}

\SetKwRepeat{Do}{do}{while}
\newcommand{\abbrev}{Multi-THuMBS}

\begin{document}

\title{\abbrev: Multi-person Tracking of \\ 3D Human Meshes Beyond Video Shots} 

\titlerunning{Multi-THuMBS}

\author{
Jeongwan On\inst{1}\and
Muhammad Salman Ali\inst{1} \and Muneeb A. Khan\inst{1}\and Sunwoo Park\inst{1}\and Inwoong Moon\inst{1} \and Hyung Jin Chang\inst{2}\and Jaekwang Kim\inst{3} \and Seong Jong Ha\inst{3}\and Seungryul Baek \inst{1}
}

\authorrunning{J.~On et al.}

\institute{UNIST, South Korea \and University of Birmingham, UK \and AI/DT Division, CJ Corporation, South Korea\\
}

\maketitle

%
\begin{center}
    \captionsetup{type=figure}
    \includegraphics[width=0.98\linewidth]{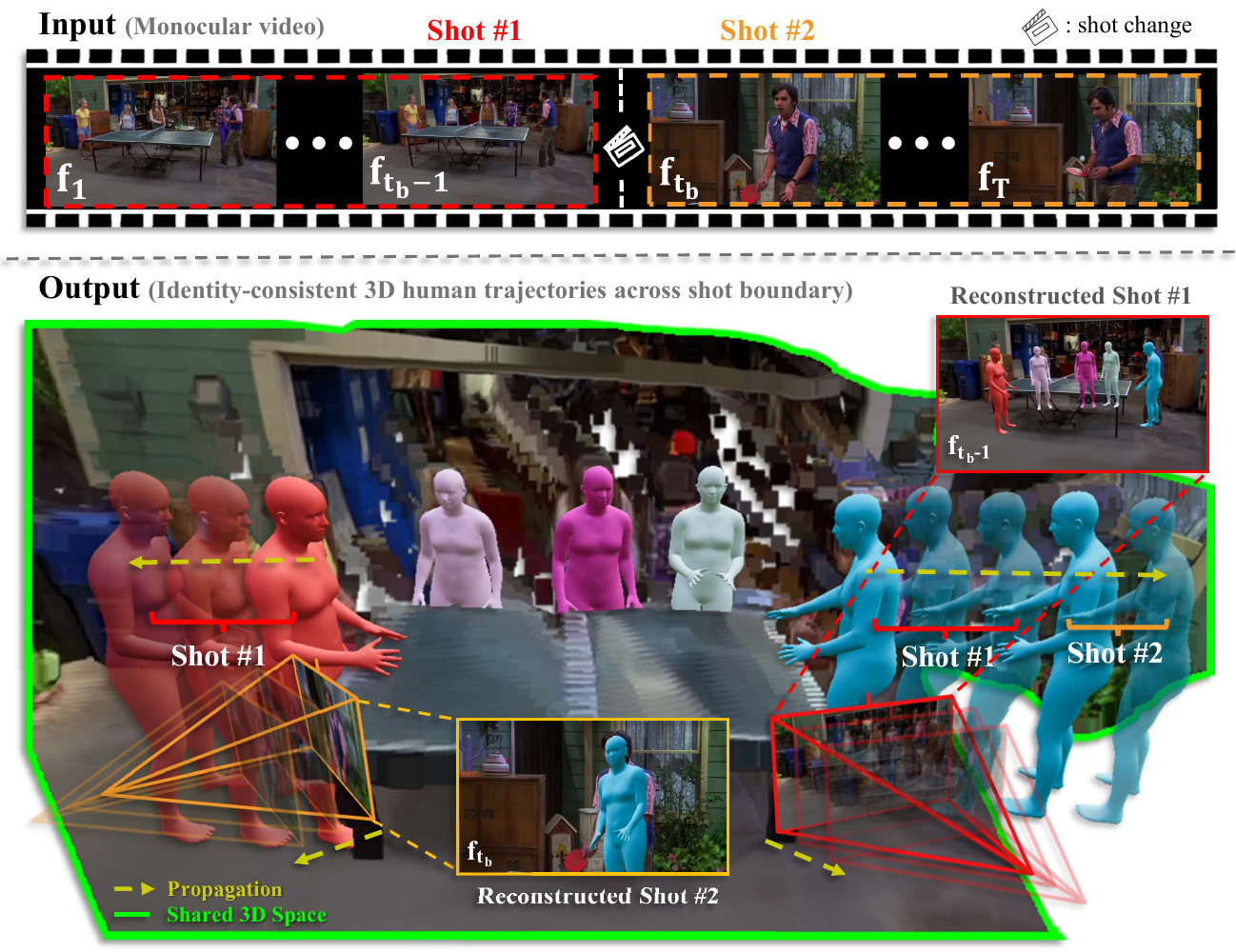}
    \captionof{figure}{Our method reconstructs globally aligned 3D human motions from multi-shot videos with multiple people, effectively handling abrupt shot changes. We apply a state-of-the-art 3D scene reconstruction technique~\cite{wang2025vggt} to two frames before and after the shot change ($\mathbf{f}_{t_b-1}$, $\mathbf{f}_{t_b}$) to construct a \underline{\textcolor{green}{Shared 3D Space}}. We further register and align the 3D human meshes of shot boundaries to the \underline{\textcolor{green}{Shared 3D Space}} and \underline{\textcolor{DYellow}{propagate}} the alignment throughout \underline{\textcolor{red}{Shot 1}} and \underline{\textcolor{orange}{Shot 2}}. Through this propagation, our method maintains consistent identity assignment and motion continuity across shot changes.}
    \label{fig:teaser}
\end{center}

\begin{abstract}
Tracking multi-person 3D human meshes from in-the-wild videos is a highly challenging problem due to complex interactions, frequent occlusions, and severe truncation inherent in unconstrained environments. While recent approaches have improved robustness against these issues, they largely overlook the critical challenge prevalent in real-world footage: frequent shot changes. These abrupt transitions in camera viewpoints often cause existing methods to lose track of human identities and fail in reconstructing temporally coherent trajectories. Although several recent works have explored 3D human mesh tracking under shot changes, they are still limited to single-person scenarios, making them inadequate for real-world videos where multiple people interact and appear simultaneously. To address this limitation, we propose \abbrev~(Multi-person Tracking of 3D Human Meshes Beyond Video Shots) that leverages a state-of-the-art 3D scene prior to reconstruct the two boundary frames in a single shared 3D space. Human meshes are then registered within the shared 3D space, maintaining per-person identity and motion consistency across shot changes. Extensive experiments demonstrate that our approach yields significant improvements in 3D human mesh recovery, camera pose estimation, and identity tracking, thereby ensuring high-fidelity motion reconstruction with consistent identity preservation across shots compared to previous state-of-the-art methods.
\end{abstract}
\section{Introduction}
\begin{figure}[t]
    \centering
    \includegraphics[width=0.98\linewidth]{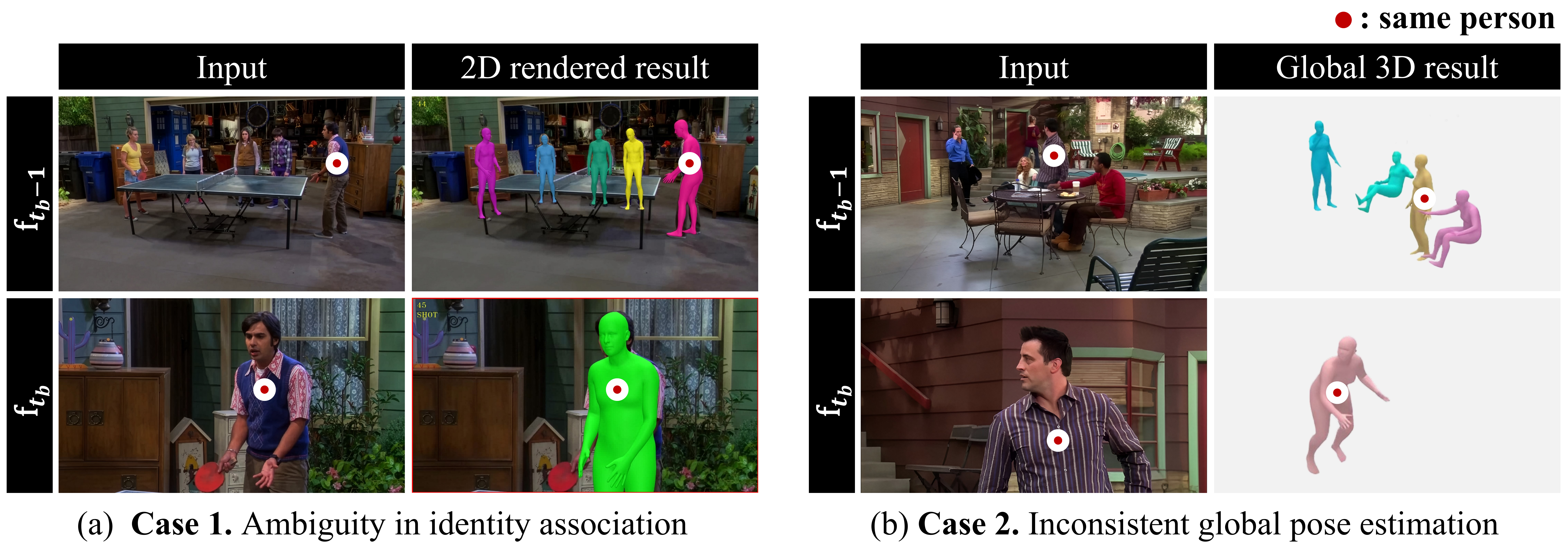}
    \caption{
    The existing state-of-the-art human mesh reconstruction method~\cite{wang2025prompthmr} suffers from two distinct limitations: (\textbf{Case 1}) First, the method incorrectly assigns different identities across the shot change, as indicated by the different mesh colors; (\textbf{Case 2}) Second, the same persons' meshes are located in the completely different 3D locations after the shot change. In both cases, the number of visible individuals varies across the shot boundary, further complicating both re-identification and global alignment.
    }
   \label{fig:fig2}
\end{figure}

Tracking and reconstruction of multi-person 3D human meshes from videos is a fundamental goal in computer vision with applications spanning animation, sports analytics, AR/VR, and human-computer interaction. Recent methods for multi-person 3D human mesh tracking~\cite{shen2024gvhmr, wang2025prompthmr, shin2023wham} have achieved remarkable accuracy in real-world videos, effectively handling challenges such as occlusion and depth ambiguity. However, these advancements heavily rely on the assumption of continuous camera motion within a single shot. Real-world videos ranging from movies and broadcast sports to edited online content are typically composed of multiple shots with abrupt transitions (as illustrated in Figures~\ref{fig:teaser} and \ref{fig:fig2}). Despite the ubiquity of such footage, the problem of tracking 3D human motion across shot changes has been largely overlooked.

To address discontinuous camera motion, multi-view reconstruction frameworks~\cite{choi2025hcp, yang2025hac, rojas2025hamst3r, mueller2024hsfm} have been proposed to integrate observations from disjoint viewpoints into a shared 3D space. However, these methods are inherently limited to simultaneous captures at single time instances, lacking the ability to model temporal human motion. Building upon these spatial alignment principles, a few pioneering works~\cite{pavlakos2022multishot, zhang2025humanmmglobalhumanmotion} have attempted to extend reconstruction across shot boundaries in multi-shot videos. However, these methods are fundamentally limited to single-person scenarios and do not address the complex identity associations required when multiple individuals appear simultaneously across different shots. As a result, although prior studies partially tackle either spatial misalignment or cross-shot reconstruction, the problem of identity-consistent multi-person 3D reconstruction across shot boundaries remains largely unsolved. This gap highlights the need for a unified framework that can jointly handle abrupt viewpoint changes, global spatial alignment, and robust multi-person identity tracking across multi-shot videos.

However, extending multi-person tracking to multi-shot scenarios introduces severe challenges that existing methods struggle to address (as shown in Fig.~\ref{fig:fig2}): (i) Abrupt shot changes introduce sudden variations in camera viewpoints, leading to drastic appearance changes in the image space. As a result, existing appearance-based re-identification methods often fail to maintain consistent identities across shots.
(ii) Shot changes also cause abrupt changes in camera positions, making it difficult to obtain globally aligned human meshes. (iii) The number of visible individuals may change across shot boundaries, making identity association ambiguous and temporal correspondence difficult to establish. In such settings, existing methods fail to recognize shot changes, leading to erroneous global pose estimates at the boundary. Moreover, by enforcing temporal continuity across these discontinuities, they introduce unrealistic motion artifacts such as foot sliding, which severely degrades reconstruction performance.

To overcome these limitations, we propose the first geometry-driven framework \abbrev~(\textbf{Multi}-person \textbf{T}racking of 3D \textbf{Hu}man \textbf{M}eshes \textbf{B}eyond Video \textbf{S}hots) designed for identity-consistent, multi-person 3D human mesh reconstruction and tracking across shot boundaries, as illustrated in Fig.~\ref{fig:teaser}. Our key insight is to leverage 3D scene geometry as a robust spatial anchor to bridge human trajectories across abrupt viewpoint changes. Given a multi-shot video, our framework first detects shot change and reconstructs 3D human meshes for each shot, along with visual cues such as human poses and UV texture maps~\cite{goel2023humans}. To establish a shared 3D space, we reconstruct dense scene point clouds and camera parameters at boundary frames by leveraging the 3D scene priors provided by VGGT~\cite{wang2025vggt}. A multi-stage optimization then aligns human meshes from adjacent shots to this shared space, maintaining spatial consistency across shot boundaries. This geometric alignment enables a re-identification (Re-ID) strategy that integrates appearance, geometry, and pose cues, rather than relying solely on appearance cues, which become unreliable due to drastic variations in camera viewpoints caused by abrupt shot changes. During Re-ID, we compute pairwise 3D distances between individuals and apply thresholding to account for the varying number of visible individuals across shot boundaries, thereby mitigating false matches. The resulting correspondences are temporally propagated throughout both shots, followed by global trajectory smoothing. At shot boundaries, a cross-camera consistency loss enforces coherent reprojections, producing globally aligned and identity-consistent 3D human trajectories across the entire video.

\noindent
In summary, our main contributions are as follows:
\vspace{-0.2cm}
\begin{itemize}
  \item We propose the \abbrev. To the best of our knowledge, this is the first work to extend multi-person 3D human mesh tracking to multi-shot settings. Our method maintains both identity and motion consistency across shot boundaries by leveraging a learned 3D prior to construct a shared 3D space for adjacent shots, within which individual trajectories are seamlessly connected.

  \item We propose a re-identification (Re-ID) strategy that integrates appearance, pose, and geometric (3D distance) cues to establish robust identity association across shot boundaries. Unlike appearance-only methods, our approach maintains reliable identity association even under drastic viewpoint shifts and severe appearance changes caused by abrupt shot changes.
  
  \item We introduce a global optimization framework that enforces spatiotemporal consistency through joint trajectory smoothing and cross-camera reprojection loss. This joint optimization stabilizes pose estimates, corrects geometric misalignment, and ensures globally consistent 3D human motion across multi-shot videos.
  
  \item Extensive experiments across multiple benchmarks encompassing multi-camera environments~\cite{EgoHumans, zhang2022egobody, Harmony4D} and dynamic in-the-wild videos~\cite{AVA, pavlakos2022sitcoms3d} demonstrate consistent improvements in pose estimation, camera tracking, and identity association. Detailed ablation studies further support the contribution of each component of our framework.

\end{itemize}
\section{Related Work}

\noindent 
\textbf{3D Human Mesh Reconstruction and Tracking.} Early research in 3D human mesh reconstruction~\cite{hmrKanazawa17, Mehta_2017} focused on single-person and single-image scenarios. While these methods established a baseline for pose estimation, they fundamentally lacked the ability to understand human motion dynamics.
%
To overcome this limitation, subsequent studies~\cite{kocabas2019vibe, mehta2018single} extended frameworks to the video domain, incorporating temporal modeling to capture motion cues. However, these approaches remained confined to single-person settings, failing to generalize to complex, in-the-wild multi-person scenes.
%
To address this, several works~\cite{rajasegaran2022tracking, goel2023humans} expanded reconstruction to multi-person tracking. Despite their success in tracking identities, they were restricted to local image-space estimation, lacking global spatial reasoning in world coordinates.
%
Addressing this, optimization-based frameworks such as SLAHMR~\cite{ye2023slahmr} emerged, jointly optimizing human and camera trajectories to achieve globally consistent reconstruction. However, their reliance on iterative optimization resulted in prohibitive inference times, making them unsuitable for scalable applications.
%
Consequently, the field shifted towards feed-forward approaches to enable more efficient global 3D human motion reconstruction. Recent methods like WHAM~\cite{shin2023wham} and PromptHMR~\cite{wang2025prompthmr} estimate global human motion using purely learning-based architectures, avoiding the high computational cost of iterative optimization.

Nevertheless, a critical limitation persists: they assume continuous camera motion and fail to handle shot changes, leading to tracking drifts or re-initialization errors at shot boundaries. Although Pavlakos \etal~\cite{pavlakos2022human} partially addressed tracking across shot changes, their scope was limited to single-person sequences. To address these challenges, HumanMM~\cite{zhang2025humanmmglobalhumanmotion} explicitly models shot changes and occlusions for multi-shot motion reconstruction, yet it remains limited to single-person scenarios. Similarly, ShowMak3r~\cite{kim2025showmak3rcompositionaltvreconstruction} jointly optimizes human pose and neural scene geometry for cross-shot alignment in multi-person settings; however, it primarily focuses on visual quality rather than motion accuracy.

To date, no prior work jointly addresses multi-person tracking, and shot discontinuity (see the Supplementary Material for a detailed comparison).

\noindent \textbf{Person Re-Identification.}
Early person re-identification (Re-ID) approaches relied on handcrafted features~\cite{farenzena2010person, gheissari2006person}, which were highly sensitive to viewpoint and illumination changes. Deep CNN-based methods~\cite{ahmed2015improved, varior2016gated} improved robustness by learning discriminative representations, while weakly supervised frameworks~\cite{liu2017end} and architecture refinements~\cite{luo2019bag} reduced dependence on large annotated datasets.
Transformer-based architectures have also advanced Re-ID performance through stronger global reasoning. ViTs~\cite{dosovitskiy2020image} enable long-range feature modeling, and TransReID~\cite{he2021transreid} adapted them for identity tracking in crowded, multi-person scenes. Pose2ID~\cite{yuan2025poses} proposed a training-free feature centralization strategy that enhances identity representation without additional Re-ID training, while KPR~\cite{somers2024keypoint} introduced keypoint-promptable matching to resolve occlusion ambiguities. 
Despite their strong identity robustness, existing Re-ID methods are primarily appearance-based and struggle at shot boundaries, where viewpoint, lighting, and appearance shifts cause unreliable identity association. Moreover, they also rely on long temporal sequences requiring high computational complexity, limiting their generalization to real-world, multi-shot scenarios.

In summary, existing approaches lack a unified framework for multi-person 3D human mesh reconstruction with motion and identity-consistent tracking across multi-shot videos. Existing state-of-the-art methods address only partial aspects of this problem: PromptHMR~\cite{wang2025prompthmr} handles multi-person tracking but is limited to single shots; HumanMM~\cite{zhang2025humanmmglobalhumanmotion} processes multi-shot videos but only for single-person; ShowMak3r~\cite{kim2025showmak3rcompositionaltvreconstruction} handles multi-person multi-shot scenarios but prioritizes rendering quality over motion and identity consistency. In contrast, our framework is the first to achieve globally consistent multi-person 3D reconstruction with motion and identity tracking across shot boundaries.

\begin{figure*}[t]
    \centering
    \includegraphics[width=\textwidth]{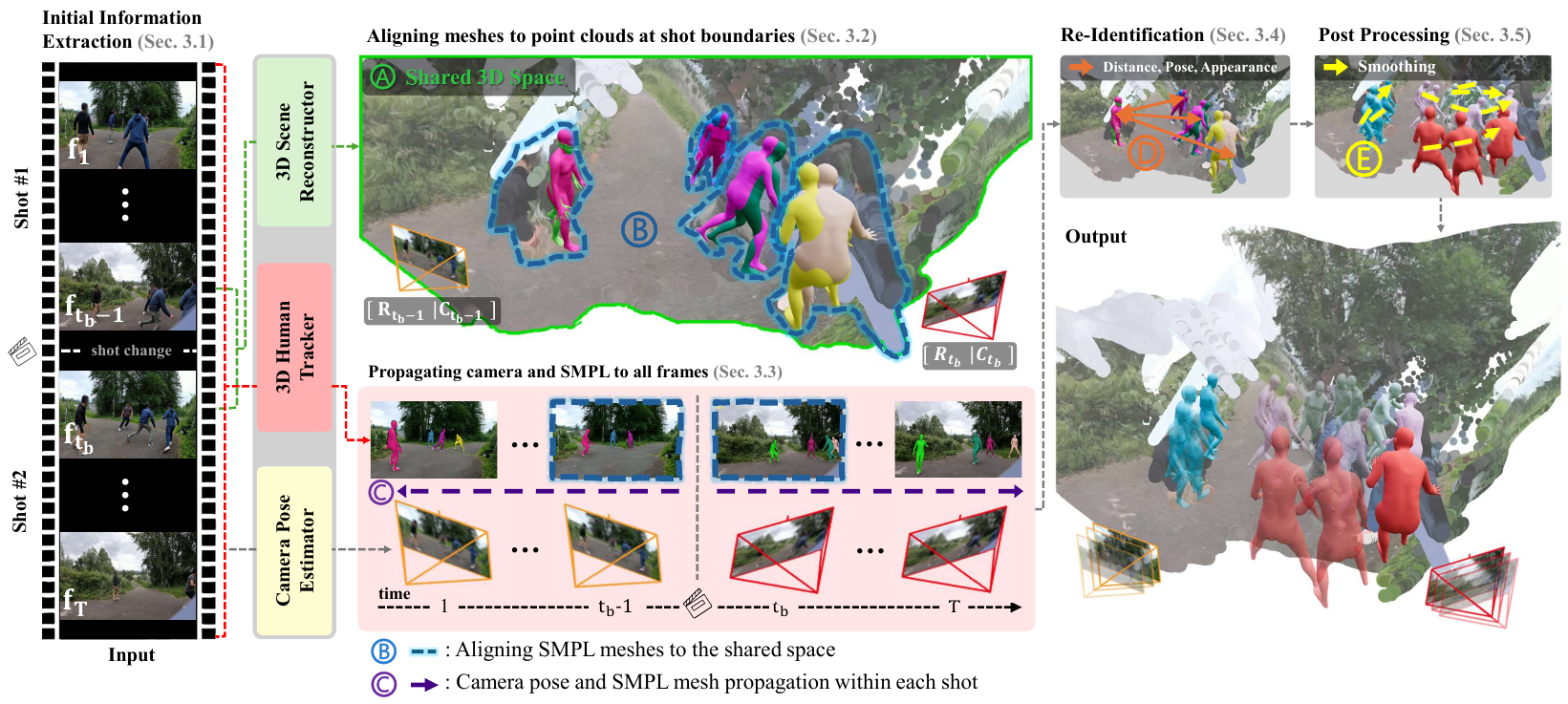}
    \caption{\textbf{Overview of the \abbrev.}
    The input video is split into two shots at the detected boundary. We estimate 3D human meshes and camera poses for all frames, reconstruct a \underline{\textcolor{green}{\textcircled{A} shared 3D space}} from the boundary frames $\{ \mathbf{f}_{t_b-1}, \mathbf{f}_{t_b} \}$, and \underline{\textcolor{Lightblue}{\textcircled{B} align}} both shots to this shared space. We then \underline{\textcolor{DDYellow}{\textcircled{C} propagate}} camera poses and global positions within each shot, \underline{\textcolor{DOrange}{\textcircled{D}link identities}} across the boundary using spatial proximity integrated with pose and appearance cues, and apply \underline{\textcolor{DYellow}{\textcircled{E} temporal smoothing}} to obtain stable, motion- and identity-consistent 3D trajectories.}
    \label{fig:fig3}
\end{figure*}

\section{Methodology}
\label{method}

Given a multi-shot video $\mathcal{V} = \{\mathcal{S}_1, \mathcal{S}_2, \ldots, \mathcal{S}_n\}$ comprising $n$ shots, we aim to reconstruct a multi-person motion and identity-consistent 3D human meshes $\mathbf{m}$ across shot boundaries. Each shot $\mathcal{S}_j$ consists of consecutive frames $\{\mathbf{f}_t\}$ captured at a specific time $t$. We focus on the two-shot case ($n=2$) for clarity; extensions to $n > 2$ are provided in supplementary material. Let $f_{t_b}$ represent the shot boundary frame, dividing $\mathcal{V}$ into $\mathcal{S}_1 = \{\mathbf{f}_t\}_{t=1}^{t_b-1}$ (pre-boundary) and $\mathcal{S}_2 = \{\mathbf{f}_t\}_{t=t_b}^T$ (post-boundary), where $T$ is the total number of frames.

We first employ PySceneDetect~\cite{pyscenedetect} to identify $t_b$, then independently reconstruct 3D human meshes $\{\mathbf{m}_t^i\}_{t=1}^T$ for each detected person $i$ using 4DHumans~\cite{goel2023humans}. To align these meshes across the boundary, we reconstruct scene point clouds $\mathbf{P}_{t_b-1}, \mathbf{P}_{t_b}$ at the frames $\mathbf{f}_{t_b-1}$,  $\mathbf{f}_{t_b}$ using VGGT~\cite{wang2025vggt}, which provides one-to-one pixel-to-point correspondence. Per-person point clouds ${\mathbf{P}_{t_b-1}^i, \mathbf{P}_{t_b}^i}$ are extracted by selecting 3D points within segmentation masks $\mathbf{M}_t^i$ obtained from Grounded SAM~\cite{ren2024grounded}. Aligning the boundary meshes $\{\mathbf{m}_{t_b-1}^i, \mathbf{m}_{t_b}^i\}$ to these point clouds synchronizes the coordinate systems of $\mathcal{S}_1$ and $\mathcal{S}_2$. 

Finally, we perform identity matching using spatial proximity fused with pose and appearance similarities and apply temporal smoothing. Each component of our framework is described in detail below, and the overall workflow is visualized in Figure~\ref{fig:fig3}.

\subsection{Initial Information Extraction}

\label{method:init}

For each shot $\mathcal{S}_1$ and $\mathcal{S}_2$, we extract 3D human meshes $\{ \mathbf{m}_t^i \}^T_{t=1}$
using 4DHumans~\cite{goel2023humans}, which provides stable intra-shot identity tracking. Each mesh $\mathbf{m}_t^i$ represents the $i$-th person at time $t$ and is parameterized by SMPL~\cite{SMPL:2015} with global orientation ${\mathbf{\Phi}}_t^i \in \mathbb{R}^{3 \times 3}$, root translation ${\mathbf{\Gamma}}_t^i \in \mathbb{R}^3$, pose parameters ${\mathbf{\theta}}_t^i \in \mathbb{R}^{23 \times 3}$, and shape parameters ${\mathbf{\beta}}^i \in \mathbb{R}^{10}$ (time-invariant). Additionally, we extract 2D keypoints $\mathbf{J}_t^i$ using ViTPose~\cite{xu2022vitpose} within the bounding boxes rendered from the 3D meshes $\mathbf{m}_t^i$, which serve as supervision for the subsequent alignment optimization.

To construct a shared 3D space across different shots, we employ VGGT~\cite{wang2025vggt} over traditional structure-from-motion (SfM) methods~\cite{wang2024dust3r, leroy2024mast3r, tang2024mv}. This is because VGGT remains robust under the sparse viewpoints and extreme viewpoint changes typically observed at shot boundaries. However, VGGT is inherently designed to process multi-view images captured at a single time step, making it infeasible to apply directly to videos. In particular, VGGT struggles with performance degradation in dynamic scenes involving moving people or backgrounds, and its high computational cost makes it unsuitable for long-term videos. Therefore, adopting the assumption from~\cite{zhang2025humanmmglobalhumanmotion} that the temporal gap at a shot boundary is negligible, we apply VGGT only to the boundary frames, $\mathbf{f}_{t_b-1}$ and $\mathbf{f}_{t_b}$, treating them as a set of multi-view captures. Specifically, we construct the 3D point clouds $\{ \mathbf{P}_{t_b-1}, \mathbf{P}_{t_b} \}$ along with their corresponding intrinsic parameters $\mathbf{K}$ and camera poses $\{ [\mathbf{R}_{t_b-1}|\mathbf{C}_{t_b-1}], [\mathbf{R}_{t_b}|\mathbf{C}_{t_b}] \}$, where $\mathbf{R}_t \in \mathbb{R}^{3 \times 3}$ and $\mathbf{C}_t \in \mathbb{R}^3$ denote the rotation and center translation, respectively. Since VGGT assigns a unique 3D point to each pixel, we then extract per-person point clouds $\{\mathbf{P}_{t_b-1}^i, \mathbf{P}_{t_b}^i\}$ by selecting points within the segmentation masks $\mathbf{M}_t^i$ obtained via Grounded SAM~\cite{ren2024grounded}.

Finally, to address the scale ambiguity between the meshes and point clouds, we compute the maximum centroid-to-vertex distance for both $\{ \mathbf{m}_{t_b}^i, \mathbf{m}_{t_b-1}^i \}$ and $\{ \mathbf{P}_{t_b}^i, \mathbf{P}_{t_b-1}^i \}$,
determine their ratio, average it across all persons, and apply this factor to uniformly rescale $\{ \mathbf{P}_{t_b}, \mathbf{P}_{t_b-1} \}$.

\subsection{Aligning Meshes to Point Clouds at Shot Boundary}
\label{method:align}

With the scale uniformly aligned, our next step is to spatially anchor the human meshes into VGGT's shared 3D space. However, the per-person 3D meshes $\{\mathbf{m}_{t_b-1}^i, \mathbf{m}_{t_b}^i\}$ and point clouds $\{\mathbf{P}_{t_b-1}^i, \mathbf{P}_{t_b}^i\}$ are initially defined in different coordinate systems. To align them, we optimize the human root translation $\{ \mathbf{\Gamma}_{t_b-1}^i, \mathbf{\Gamma}_{t_b}^i \}$, global orientation $\{ \mathbf{\Phi}_{t_b-1}^i, \mathbf{\Phi}_{t_b}^i \}$, and the camera pose $\{ [\mathbf{R}_{t_b-1}|\mathbf{C}_{t_b-1}]$, $[\mathbf{R}_{t_b}|\mathbf{C}_{t_b}] \}$. Direct joint optimization of these spatial parameters, however, is highly non-convex and often becomes unstable due to severe parameter coupling. To address this, we propose a three-stage progressive optimization (Algorithm~\ref{alg:align}) that sequentially refines them. The full optimization objective combines 2D reprojection, silhouette alignment, and depth consistency:
\begin{equation}
    \min_{\mathbf{\Gamma}_t^i, \mathbf{\Phi}_t^i, [\mathbf{R}_t|\mathbf{C}_t]} 
    \lambda_{\text{2D}} \mathcal{L}_{\text{2D}} + \lambda_{\text{sil}} \mathcal{L}_{\text{sil}} + \lambda_{\text{depth}} \mathcal{L}_{\text{depth}},
    \label{eq:full_objective}
\end{equation}
where $t \in \{{t_b}-1, t_b\}$, $\mathcal{L}_{\text{2D}}$ enforces 2D reprojection consistency, $\mathcal{L}_{\text{sil}}$ ensures silhouette consistency, and $\mathcal{L}_{\text{depth}}$ enforces depth alignment with scene geometry. These losses and parameters are optimized across three stages to avoid local minima, as detailed below.

In the first stage, we initialize the root translation $\{ \mathbf{\Gamma}_{t_b-1}^i, \mathbf{\Gamma}_{t_b}^i \}$ by leveraging the one-to-one mapping between the image pixels and the VGGT point cloud. Specifically, we extract the 2D root joint from the detected keypoints $\{ \mathbf{J}_{t_b-1}^i, \mathbf{J}_{t_b}^i \}$ and assign the 3D coordinate of its corresponding VGGT point as the initial translation.

\begin{wrapfigure}{l}{0.6\linewidth}
\vspace{-1.5cm}
\centering
\begin{minipage}{\linewidth}
    \begin{algorithm}[H]
    \small
    \caption{Aligning meshes to point clouds at shot boundary}
    \label{alg:align}
    \KwIn{$\{\mathbf{m}_{t_b-1}^i, \mathbf{m}_{t_b}^i\}$: Per-person local meshes}
    \KwIn{$[\mathbf{R}_{t_b-1}|\mathbf{C}_{t_b-1}], [\mathbf{R}_{t_b}|\mathbf{C}_{t_b}]$: Camera poses}
    \KwOut{$\{{}^w\mathbf{m}_{t_b-1}^i, {}^w\mathbf{m}_{t_b}^i\}$: Global meshes}
    \KwOut{$[\hat{\mathbf{R}}_{t_b-1}|\hat{\mathbf{C}}_{t_b-1}], 
    [\hat{\mathbf{R}}_{t_b}|\hat{\mathbf{C}}_{t_b}]$: Refined poses}
    \hrulefill\\
    $\text{max\_iter} \leftarrow [500, 500, 1500]$; \\
    \For{stage $s = 1$ \KwTo $3$}{
        $\text{iter} \leftarrow 1$; \\
        \Repeat{$\text{iter} > \text{max\_iter}[s]$}{
            \For{$t \in \{t_b-1, t_b\}$}{
                \For{each person $i$}{
                    \uIf{$s = 1$}{
                        \, \textbf{Initialize} $\mathbf{\Gamma}_t^i$; \\
                    }
                    \uElseIf{$s = 2$}{
                        \, \textbf{Optimize} $\mathbf{\Gamma}_t^i, \mathbf{\Phi}_t^i$; \\
                        \textbf{Minimize} $\lambda_{\text{2D}}^{(2)} \mathcal{L}_{\text{2D}}$; \\
                    }
                    \uElse{
                        \, \textbf{Optimize} $\mathbf{\Gamma}_t^i, \mathbf{\Phi}_t^i, 
                        [\mathbf{R}_t|\mathbf{C}_t]$; \\
                        \textbf{Minimize} $\lambda_{\text{2D}}^{(3)} \mathcal{L}_{\text{2D}} + \lambda_{\text{sil}} \mathcal{L}_{\text{sil}} + \lambda_{\text{depth}} \mathcal{L}_{\text{depth}}$;
                    }
                }
            }
            $\text{iter} \leftarrow \text{iter} + 1$;
        }
    }
    \end{algorithm}
\end{minipage}
\vspace{-2cm}
\end{wrapfigure}

In the second stage, we jointly refine both
$\{ \mathbf{\Gamma}_{t_b-1}^i, \mathbf{\Gamma}_{t_b}^i \}$ and $\{ \mathbf{\Phi}_{t_b-1}^i, \mathbf{\Phi}_{t_b}^i \}$ for $500$ iterations
to better align with 2D observations with $\lambda_{\text{2D}}^{(2)} = 10.0$:
\begin{equation}
    \min_{\mathbf{\Gamma}_t^i, \mathbf{\Phi}_t^i} \lambda_{\text{2D}}^{(2)} \mathcal{L}_{\text{2D}}.
    \label{eq:stage2}
\end{equation}
The 2D reprojection loss $\mathcal{L}_{\text{2D}}$ enforces consistency between projected 3D joints and detected 2D keypoints:
\vspace{0.25cm}
\begin{align}
    \nonumber \mathcal{L}_{\text{2D}} = \sum_i \| &\Pi_f(\mathbf{R}_t \cdot J(\phi_t^i) \\
    &+ \mathbf{C}_t) - \mathbf{J}_t^i \|_2^2,
    \label{eq:loss_2d}
\end{align}
\noindent where $\phi^i_t$ denotes SMPL parameters (i.e., $\mathbf{\Gamma}^i_t$, $\mathbf{\Phi}^i_t$, $\theta^i_t$, $\beta^i_t$) and $\mathbf{J}_t^i$ are 2D keypoints extracted using ViTPose~\cite{xu2022vitpose}, $J(\cdot)$ denotes the SMPL joint regressor~\cite{SMPL:2015}, and $\Pi_f(\cdot)$ represents perspective projection using camera intrinsics $\mathbf{K} \in \mathbb{R}^{3\times3}$.

\vspace{0.5cm}
In the final stage, we incorporate 3D geometric scene constraints to resolve depth ambiguity by optimizing all parameters in Eq.~\ref{eq:full_objective} for 1500 iterations with 
$\lambda^{(3)}_{\text{2D}} = 50.0$, $\lambda_{\text{sil}} = 0.01$, and $\lambda_{\text{depth}} = 1.0$.

For each boundary frame $t \in \{t_b-1, t_b\}$, the depth consistency loss $\mathcal{L}_{\text{depth}}$ penalizes the 3D distance between SMPL vertices $\mathbf{v}_k$ whose projections lie within $\mathbf{M}_t^i$ and corresponding scene points $\mathbf{p}_k \in \mathbf{P}_t^i$:
\begin{equation}
    \mathcal{L}_{\text{depth}} = \sum_k \|\mathbf{v}_k - \mathbf{p}_k\|_2^2.
    \label{eq:loss_depth}
\end{equation}

Similarly, the silhouette loss $\mathcal{L}_{\text{sil}}$ penalizes vertices projecting outside $\mathbf{M}_t^i$:
\begin{equation}
    \mathcal{L}_{\text{sil}} = \sum_{\mathbf{v}_k} 
    \text{dist}\left(\Pi_f(\mathbf{v}_k)\right),
    \label{eq:loss_sil}
\end{equation}
where $\text{dist}(u) \in \mathbb{R}^{H \times W}$ is a precomputed distance transform assigning each pixel $u$ its distance to the nearest mask boundary~\cite{distance}.

This process produces globally aligned meshes $\{{}^w\mathbf{m}_{t_b-1}^i, {}^w\mathbf{m}_{t_b}^i\}$, derived from the aligned SMPL parameters $\{ {}^w\mathbf{\Phi}_t, {}^w\mathbf{\Gamma}_t, \theta_t, \beta \}_{t \in \{t_b-1, t_b\}}$, along with refined camera poses $\{ [\hat{\mathbf{R}}_{t_b-1}|\hat{\mathbf{C}}_{t_b-1}], [\hat{\mathbf{R}}_{t_b}|\hat{\mathbf{C}}_{t_b}] \}$. These outputs serve as anchors for subsequent temporal propagation.

\subsection{Propagating Camera Poses and SMPL Parameters}
\label{method:propagate}

As discussed in Sec.~\ref{method:init}, while VGGT~\cite{wang2025vggt} successfully constructs a shared 3D space at the boundary frames $\mathbf{f}_{t_b-1}$ and $\mathbf{f}_{t_b}$, applying it to all video frames is computationally prohibitive and prone to errors in dynamic scenes with significant human and background motion. To efficiently track cameras across non-boundary frames, we adopt DROID-SLAM~\cite{teed2021droid}, which provides lightweight, robust, and temporally smooth camera tracking within a single continuous shot. Therefore, to achieve global consistency across multiple shots, we align these intra-shot trajectories to our shared global space by computing the relative transformation between the DROID-SLAM and VGGT camera poses at the shot boundary.

Specifically, for each boundary frame $t \in \{ t_b-1, t_b \}$, the relative transformation matrix $\mathcal{R}_{t} \in SE(3)$ is computed as:
\begin{equation}
\mathcal{R}_{t} = [\hat{\mathbf{R}}_{t}|\hat{\mathbf{C}}_{t}] \cdot 
[\mathbf{R}^{*}_{t}|\mathbf{C}^{*}_{t}]^{-1},
\end{equation}
\noindent where $[\mathbf{R}^{*}_{t}|\mathbf{C}^{*}_{t}]$ denote the camera poses estimated by DROID-SLAM. Here, $\mathcal{R}_{t}$ serves as the transformation aligning the local DROID-SLAM coordinate system to our shared global space. Once computed at the boundary, these transformations are propagated to all non-boundary frames within their respective shots. Specifically, $\mathcal{R}_{t_b-1}$ is applied to the DROID-SLAM poses $[\mathbf{R}^{*}_{t}|\mathbf{C}^{*}_{t}]$ for all frames where $t \in \mathcal{S}_1$, and similarly, $\mathcal{R}_{t_b}$ is applied to the cameras for all frames where $t \in \mathcal{S}_2$. Through this propagation, the locally consistent intra-shot trajectories are seamlessly integrated into a unified global coordinate system.

Similarly, we propagate the globally aligned SMPL parameters, specifically the global orientation ${}^w\mathbf{\Phi}_t^i$ and root translation ${}^w\mathbf{\Gamma}_t^i$, from the boundary frames to all non-boundary frames. This step is required because only the human meshes at the boundary frames are aligned to the shared global space. By applying corresponding relative transformations to the SMPL parameters, we ensure that the entire sequence of human meshes is temporally coherent and accurately situated within the shared global space. The detailed algorithm for propagation is provided in the supplementary material.

\subsection{Re-Identification}
\label{method:reid}

With aligned meshes in a shared 3D space, identity matching fundamentally reduces to spatial proximity. For boundary frames $\mathbf{f}_{t_b-1}$ and $\mathbf{f}_{t_b}$, we establish identity correspondences by computing pairwise Euclidean distances between root translations ${}^w\mathbf{\Gamma}^i_{t_b-1}$ and ${}^w\mathbf{\Gamma}^j_{t_b}$ of globally aligned meshes ${}^w\mathbf{m}_{t_b-1}^i$ and ${}^w\mathbf{m}_{t_b}^j$:
\begin{equation}
    \mathbb{D}_{ij} = \left\| {}^w\mathbf{\Gamma}_{t_b-1}^i - {}^w\mathbf{\Gamma}_{t_b}^j \right\|_2,
    \label{distanceterm}
\end{equation}
\noindent where $i$ and $j$ denote index of each mesh in the pre- and post-boundary frames.

To further strengthen identity association under challenging viewpoints and lighting variations, we incorporate complementary appearance and pose cues. Appearance dissimilarity $\mathbb{A}_{ij}$ compares UV texture maps extracted using 4DHumans~\cite{goel2023humans}, computing pixel-wise color differences over valid non-occluded regions. Pose dissimilarity $\mathbb{P}_{ij}$ measures axis-angle differences between SMPL joint rotations, considering only joints visible in both frames based on projected 2D keypoint visibility. These cues are integrated into a unified cost matrix:
\begin{equation}
    \mathbb{U}_{ij} = \lambda_{geo} \mathbb{D}_{ij} + \lambda_{app} \mathbb{A}_{ij} + \lambda_{pos} \mathbb{P}_{ij},
\end{equation}
\noindent 
where $\lambda_{geo}=1.0$, $\lambda_{app}=0.2$, and $\lambda_{pos}=0.35$ denote the weights for the geometric, appearance, and pose terms, respectively. An ablation study analyzing the impact of these hyperparameters is provided in the Supplementary Material. We then apply the Hungarian algorithm~\cite{Kuhn1955Hungarian} on $\mathbb{U}_{ij}$ to obtain the optimal one-to-one matching. To prevent false associations when individuals enter or exit the scene, we discard any matched pairs with a spatial distance $\mathbb{D}_{ij} > \tau$, where $\tau = 1.0$ meter. This hybrid formulation improves the geometry-driven matching and identity disambiguation in cases where spatial proximity alone is insufficient. Ultimately, this process yields the final set of matched human pairs $\mathcal{M}$ across the shot boundary.

\subsection{Post-processing}
\label{method:post}

Finally, we apply a post-processing step to refine temporal and geometric consistency across the entire video. This step smooths joint trajectories, maintains plausible body poses, and enforces cross-camera geometric consistency at the boundaries, yielding stable multi-person motions. Specifically, we formulate an optimization objective that integrates a joint-level temporal smoothing term $\mathcal{L}_\text{smooth}$ across all frames with a cross-camera reprojection constraint $\mathcal{L}_\text{cross}$ at the shot boundaries:
\begin{equation}
        \underset{{}^{w}\phi_t^i}{\min}\;
        \lambda_\text{smooth}\,\mathcal{L}_\text{smooth} +
        \lambda_\text{cross}\,\mathcal{L}_\text{cross},
    \label{eq:post}
\end{equation}
\noindent where ${}^w\phi_{t}^i$ denotes the set of globally aligned SMPL parameters (\ie, ${}^w\mathbf{\Phi}^i_{t}, {}^w\mathbf{\Gamma}^i_{t}, \theta^i_{t},$ $\beta^i_{t}$), and the loss weights are set to $\lambda_{\text{smooth}} = 10.0$ and $\lambda_{\text{cross}} = 1.0$.

The temporal smoothing loss $\mathcal{L}_\text{smooth}$ enforces temporal continuity of joint trajectories while regularizing body shape and pose parameters to maintain plausible human configurations:
\begin{equation}
\mathcal{L}_\text{smooth} =  \sum_{t}\sum_{i} \left\| {}^w\mathbf{J}_t^i - {}^w\mathbf{J}_{t+1}^i \right\|_2^2+ \sum_i \left\| \beta^i \right\|_2^2+ \sum_{i,t} \left\| \zeta_t^i \right\|_2^2,
\end{equation}
\noindent where ${}^w\mathbf{J}^i_t$ denotes 3D joints of globally aligned mesh, $\beta^i$ denotes the SMPL shape parameter, and $\zeta_t^i$ denotes the VPoser~\cite{SMPL-X:2019} latent vector representing pose ${}^w\theta_t^i$.
The cross-camera loss $\mathcal{L}_\text{cross}$ extends the 2D reprojection loss (Eq.~\ref{eq:loss_2d}) to enforce geometric consistency across two different camera viewpoints at the shot boundary. For a matched person pair $i \in \mathcal{M}$, the mesh from one shot is projected using the camera of the adjacent shot, and vice versa:
\begin{align}
\mathcal{L}_\text{cross} &= 
\sum_{i \in \mathcal{M}} \Big[
\left\| \Pi_f\left(\mathbf{R}_{t_b} J({}^w\phi_{t_b-1}^i) + \mathbf{C}_{t_b}\right)
    - \mathbf{J}_{t_b}^i \right\|_2^2 \nonumber \\
+ &\left\| \Pi_f\left(\mathbf{R}_{t_b-1} J({}^w\phi_{t_b}^i) + \mathbf{C}_{t_b-1}\right)
    - \mathbf{J}_{t_b-1}^i \right\|_2^2
\Big],
\end{align}
By refining the pose trajectories and ensuring coherent reprojections across adjacent shots, the post-processing stage effectively produces stable, identity-consistent 3D human motions throughout the entire multi-shot sequence.
\section{Experiments}

\subsection{Implementation Details}
\label{sec:exp_dataset}
Our framework is implemented in PyTorch~\cite{paszke2019pytorch} and optimized using the AdamW~\cite{loshchilov2017decoupled} optimizer. For a video containing 150 frames at 1920$\times$1080 resolution, the complete optimization takes approximately 10 minutes on a single NVIDIA RTX 3090 GPU. All detailed dataset constructions, implementation settings, and evaluation protocols are provided in the supplementary material.

\noindent \textbf{Datasets.} 
To comprehensively evaluate our method, we construct a multi-shot benchmark by modifying multi-view sequences from EgoHumans~\cite{EgoHumans}, EgoBody~\cite{zhang2022egobody}, and Harmony4D~\cite{Harmony4D} to introduce shot boundaries at specific frames. For qualitative and quantitative motion evaluation, we use AVA~\cite{AVA} and sitcom videos from \textit{Friends} (1994) and \textit{The Big Bang Theory} (2007), which naturally contain shot changes. All videos are segmented into clips based on shot boundaries to align with our evaluation protocol.

\noindent \textbf{Metrics.}
We evaluate pose with MPJPE and MPVPE, temporal smoothness with Accel, camera localization with ATE, and identity consistency via identity switches (IDs). We report three MPJPE variants: W-MPJPE (initial-frame alignment) for trajectory consistency, and WA-MPJPE (trajectory-level alignment) for shape accuracy. For real videos lacking ground-truth annotations, we additionally evaluate cross-shot motion quality using \textit{cross-shot} PCK (PCK$^*$)~\cite{pavlakos2022multishot}, Jitter, and Foot Sliding (FS).


\begin{table*}[t]
\parbox{.597\linewidth}{
    \centering
\caption{Quantitative comparison with state-of-the-art methods on EgoHumans, EgoBody, and Harmony4D. We report human pose metrics (W-MPJPE, WA-MPJPE, MPJPE, MPVPE, Acceleration).}
\label{table:table1}
\resizebox{0.98\linewidth}{!}{%
\begin{tabular}{cl|ccccc}
\hline
\multirow{2}{*}{} & \multicolumn{1}{c}{\multirow{2}{*}{\textbf{Method}}} & \multicolumn{5}{c}{\textbf{Human Metrics}} \\
& \multicolumn{1}{c}{}  & W-MPJPE$\downarrow$    &  WA-MPJPE$\downarrow$    &  MPJPE$\downarrow$    &  MPVPE$\downarrow$    &  Accel$\downarrow$          \\ \hline

\multirow{5}{*}{\rotatebox[origin=c]{90}{\color{olive} EgoHumans}}
& \multicolumn{1}{l|}{Multishot~\cite{pavlakos2022multishot}} & 474.1 & 287.4 & 347.1 & 408.2 & 63.15 \\
& \multicolumn{1}{l|}{GVHMR~\cite{shen2024gvhmr}} & 404.8 & 204.7 & 287.4 & 371.4 & 59.0 \\
& \multicolumn{1}{l|}{PromptHMR~\cite{wang2025prompthmr}} & 1778.2 & 440.9 & 285.3 & 364.1 & 74.3  \\
& \multicolumn{1}{l|}{HSfM$^\dagger$~\cite{mueller2024hsfm}} & 544.2 & 187.7 & 263.1 & 294.3 & 48.7  \\
& \multicolumn{1}{l|}{\graycb{Ours}} & \graycb{279.0} & \graycb{166.0}  & \graycb{228.3} & \graycb{262.2} & \graycb{27.3}  \\ 
\hline
\multirow{5}{*}{\rotatebox[origin=c]{90}{\color{olive} EgoBody}}
& \multicolumn{1}{l|}{Multishot~\cite{pavlakos2022multishot}} & 185.1 & 144.1 & 147.1 & 166.5 & 17.9 \\
& \multicolumn{1}{l|}{GVHMR~\cite{shen2024gvhmr}} & 174.0 & 133.3 & 108.0 & 147.1 & 14.7 \\
& \multicolumn{1}{l|}{PromptHMR~\cite{wang2025prompthmr}} & 1228.1 & 395.3 & 99.0 & 133.9 & 17.4  \\
& \multicolumn{1}{l|}{HSfM$^\dagger$~\cite{mueller2024hsfm}} & 113.1 & 96.3 & 113.3 & 123.2 & 10.9 \\
& \multicolumn{1}{l|}{\graycb{Ours}} & \graycb{99.2} & \graycb{72.8} & \graycb{72.0} & \graycb{94.9} & \graycb{6.0} \\ 
            
\hline

\multirow{5}{*}{\rotatebox[origin=c]{90}{\color{olive} Harmony4D}} 
& \multicolumn{1}{l|}{Multishot~\cite{pavlakos2022multishot}} & 248.0 & 231.6 & 511.2 & 609.5 & 37.1  \\
& \multicolumn{1}{l|}{GVHMR~\cite{shen2024gvhmr}}             & 244.9 & 166.7 & 244.7 & 334.1 & 29.6  \\
& \multicolumn{1}{l|}{PromptHMR~\cite{wang2025prompthmr}}     & 1746.3 & 399.8 & 675.7 & 746.0 & 66.8 \\
& \multicolumn{1}{l|}{HSfM$^\dagger$~\cite{mueller2024hsfm}}  & 372.0 & 178.4 & 225.6 & 257.6 & 28.3  \\
& \multicolumn{1}{l|}{\graycb{Ours}}  & \graycb{221.0} & \graycb{116.9}  & \graycb{215.9} & \graycb{278.3} & \graycb{17.4} \\
\hline
\end{tabular}%
}
}
\hfill
\hbox{\parbox{.383\linewidth}{
    \centering
\caption{Comparison of cross-shot Re-ID (IDs) and camera pose estimation (ATE) on EgoHumans, EgoBody, and Harmony4D. Ours (\textit{dist.}) denotes the baseline that uses only 3D distance term (Eq.~\ref{distanceterm}) for the Re-ID.}
\label{table:table2}
\resizebox{0.98\linewidth}{!}{%
    \begin{tabular}{c|cccc|}
        \toprule
        \multicolumn{1}{c}{\multirow{2}{*}{\textbf{Method}}} &
        \multicolumn{3}{c}{\textbf{Re-ID Metrics (IDs$\downarrow$)}} \\
        & {\color{olive} EgoHumans} & {\color{olive} EgoBody} & {\color{olive} Harmony4D}    \\
        \midrule
        PromptHMR~\cite{wang2025prompthmr} & 10.40 & 1.29 & 8.00   \\
        HSfM$^\dagger$~\cite{mueller2024hsfm} & 3.87 & 0.20 & 1.58  \\
        KPR~\cite{somers2024keypoint} & 2.54 & 0.05 & 1.19    \\
        Pose2ID~\cite{yuan2025poses} & 4.62 & 0.72 & 1.32    \\
        \midrule
        {Ours (\textit{dist.})} & {1.66} & {0.00} & {0.54}  \\ 
        \graycb{Ours} & \graycb{0.97} & \graycb{0.00} & \graycb{0.46} \\
        \bottomrule
    \end{tabular}
}
    \centering
\resizebox{0.98\linewidth}{!}{%
\begin{tabular}{cccc}
\toprule
\multirow{2}{*}{\textbf{Method}} & \multicolumn{3}{c}{\textbf{Camera Metrics (ATE$\downarrow$)}} \\
& \multicolumn{1}{c}{\color{olive} EgoHumans} & \multicolumn{1}{c}{\color{olive} EgoBody} & \multicolumn{1}{c}{\color{olive} Harmony4D}    \\ \hline
\multicolumn{1}{l|}{VGGT~\cite{wang2025vggt}}        & 1.4           & 1.3          & 1.4          \\
\multicolumn{1}{l|}{PromptHMR~\cite{wang2025prompthmr}}   & 2.3           & 1.6          & 2.3          \\
\multicolumn{1}{l|}{HSfM$^\dagger$~\cite{mueller2024hsfm}}        & 2.8           & 2.9          & 3.2          \\ \hline
\multicolumn{1}{l|}{\graycb{Ours}}        & \graycb{0.7}           & \graycb{0.1}          & \graycb{0.7}          \\ 
\bottomrule
\end{tabular}%
}
}}
\end{table*}

\begin{table*}[t]
\parbox{.597\linewidth}{
    \centering
    \caption{\textbf{Ablation study on EgoHumans.} (1) `\textit{w/o cam}' denotes a baseline that removes camera pose optimization. (2) `\textit{w/o post}' removes post-processing in Sec.~\ref{method:post}. (3) `\textit{w/o stage}' disables the hierarchical optimization in Sec.~\ref{method:align}.
    (4) `\textit{w/o align}' skips the alignment in Sec.~\ref{method:align} entirely.
    }
    \label{table:table3}    
    \resizebox{0.98\linewidth}{!}{%
        \begin{tabular}{l|cccc|c}
            \toprule
            \multicolumn{1}{c}{\multirow{2}{*}{\textbf{Method}}} &
            \multicolumn{4}{c}{\textbf{Human Metrics}} &
            \multicolumn{1}{c}{\textbf{Camera Metrics}} \\
            & W-MPJPE$\downarrow$ & MPJPE$\downarrow$ & MPVPE$\downarrow$ & Accel$\downarrow$ & ATE$\downarrow$ \\
            \midrule
            (1) w/o \textit{cam} & 389.7 & 230.1 & 264.7 & 33.7 & 1.40 \\
            (2) w/o \textit{post} & 311.2 & 241.6 & 298.3 & 47.9 & 0.77 \\
            (3) w/o \textit{stage} & 491.9 & 368.1 & 359.4 & 33.9 & 2.75 \\
            (4) w/o \textit{align} & 882.7 & 422.4 & 392.1 & 34.9 & 1.40 \\
            \midrule
            \graycb{Ours} & \graycb{278.8} & \graycb{228.3} & \graycb{262.1} & \graycb{27.3} & \graycb{0.77} \\
            \bottomrule
        \end{tabular}
    }
}
\hfill
\parbox{.383\linewidth}{
    \centering
\caption{Quantitative comparison with the state-of-the-art method on edited videos (\ie, AVA, \textit{Friends}, and \textit{The Big Bang Theory}). We report motion quality metrics (PCK$^*$, Jitter, FS). PCK$^*$ denotes \textit{cross-shot} PCK~\cite{pavlakos2022multishot}}
\label{table:table4}
\resizebox{0.98\linewidth}{!}{%
\begin{tabular}{c|ccc}
    \toprule
    \multicolumn{1}{c}{
        \multirow{2}{*}{\textbf{Method}}
    } &
    \multicolumn{3}{c}{\textbf{Motion quality}} \\
    & \bl PCK$^{*}\uparrow$ & \bl Jitter$\downarrow$ \bl & \bl $\;$FS$\downarrow$ \bl \\
\midrule

PromptHMR~\cite{wang2025prompthmr} & 62.7 & 162.44 & 23.11 \\
\midrule

\graycb{Ours} & \graycb{90.7} & \graycb{31.5} & \graycb{10.7} \\
\bottomrule
\end{tabular}%
}
}
\end{table*}


\subsection{Results}
\label{sec:exp_results}

\noindent \textbf{Baselines.}
We compare our approach against state-of-the-art methods for multi-human 3D reconstruction and tracking including Multishot~\cite{pavlakos2022multishot}, GVHMR~\cite{shen2024gvhmr}, PromptHMR~\cite{wang2025prompthmr}, and HSfM$^\dagger$. where HSfM$^\dagger$ denotes a modified variant of HSfM~\cite{mueller2024hsfm} adapted to our multi-shot, multi-person setting (implementation details in supplementary material).

For camera pose estimation, we additionally evaluate against PromptHMR~\cite{wang2025prompthmr}, VGGT~\cite{wang2025vggt}, and HSfM$^\dagger$, which provide comparisons for our boundary-aware camera optimization approach. For re-identification across shot boundaries, we include appearance-based methods KPR~\cite{somers2024keypoint} and Pose2ID~\cite{yuan2025poses}, which rely on visual features and are sensitive to viewpoint and illumination changes.

\begin{figure}[t]
    \centering
    \includegraphics[width=1.0\linewidth]{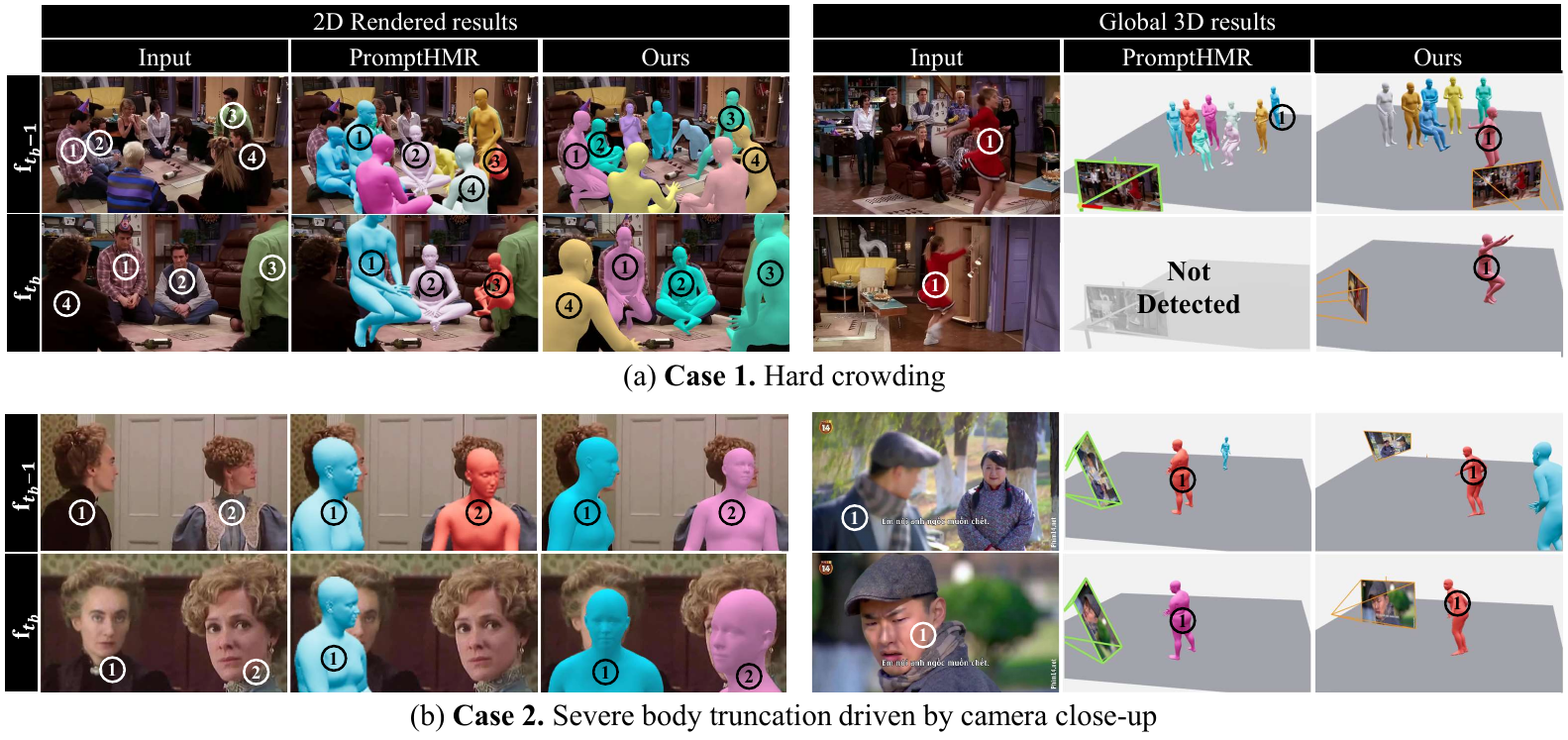}
    \caption{
    \textbf{Qualitative results for challenging cases:} (\textbf{Case 1}) When the number of visible persons changes drastically across a shot boundary, PromptHMR loses track of specific individuals, whereas our method successfully maintains consistent tracking.  (\textbf{Case 2}) Under severe body truncation driven by camera close-up, PromptHMR predicts completely incorrect camera poses, while our method robustly estimates accurate camera locations and global trajectories.
    }
   \label{fig:fig4}
\end{figure}

\noindent \textbf{Quantitative results.} {%
As shown in Table~\ref{table:table1}, our method achieves state-of-the-art spatial pose accuracy and temporal smoothness on EgoHumans~\cite{EgoHumans}, EgoBody~\cite{zhang2022egobody}, and Harmony4D~\cite{Harmony4D}. For camera estimation and Re-ID (Table~\ref{table:table2}), we achieve the lowest ATE, demonstrating robust alignment under abrupt viewpoint changes. For Re-ID, both our spatial-only and full variants substantially outperform appearance-based baselines~\cite{somers2024keypoint, yuan2025poses}, validating the efficacy of our shared 3D space. Finally, on in-the-wild datasets (AVA, sitcoms), our method significantly outperforms PromptHMR~\cite{wang2025prompthmr} in PCK$^*$, Jitter, and FS (Table~\ref{table:table4}), proving our ability to generate smooth, physically plausible transitions across abrupt shot changes.

\noindent \textbf{Qualitative results.} Figures~\ref{fig:fig4}--\ref{fig:fig5} qualitatively compare our pipeline with Prompt-HMR~\cite{wang2025prompthmr}. Figure~\ref{fig:fig4} demonstrates our superior camera pose estimation and person tracking at shot boundaries. Under challenging conditions like hard crowding and severe occlusion, PromptHMR frequently loses track of specific individuals or predicts completely incorrect camera poses. In contrast, our method reliably re-identifies persons across transitions and recovers accurate camera parameters, ensuring correct mesh-image alignment.

Furthermore, our method effectively suppresses temporal drift in global 3D trajectories (Fig.~\ref{fig:fig5}). PromptHMR exhibits noticeable sliding artifacts and abrupt positional shifts at shot boundaries, particularly during rapid viewpoint changes or zoom-ins in datasets like EgoHumans~\cite{EgoHumans} and AVA~\cite{AVA}. Conversely, our approach maintains stable and coherent multi-person motion trajectories across these complex transitions.

Overall, these improvements underline the robustness of our pipeline in handling challenging shot transitions, achieving temporally consistent and spatially accurate reconstructions where prior methods struggle.

\begin{wrapfigure}{l}{0.6\textwidth}
    \centering
    \includegraphics[width=0.99\linewidth]{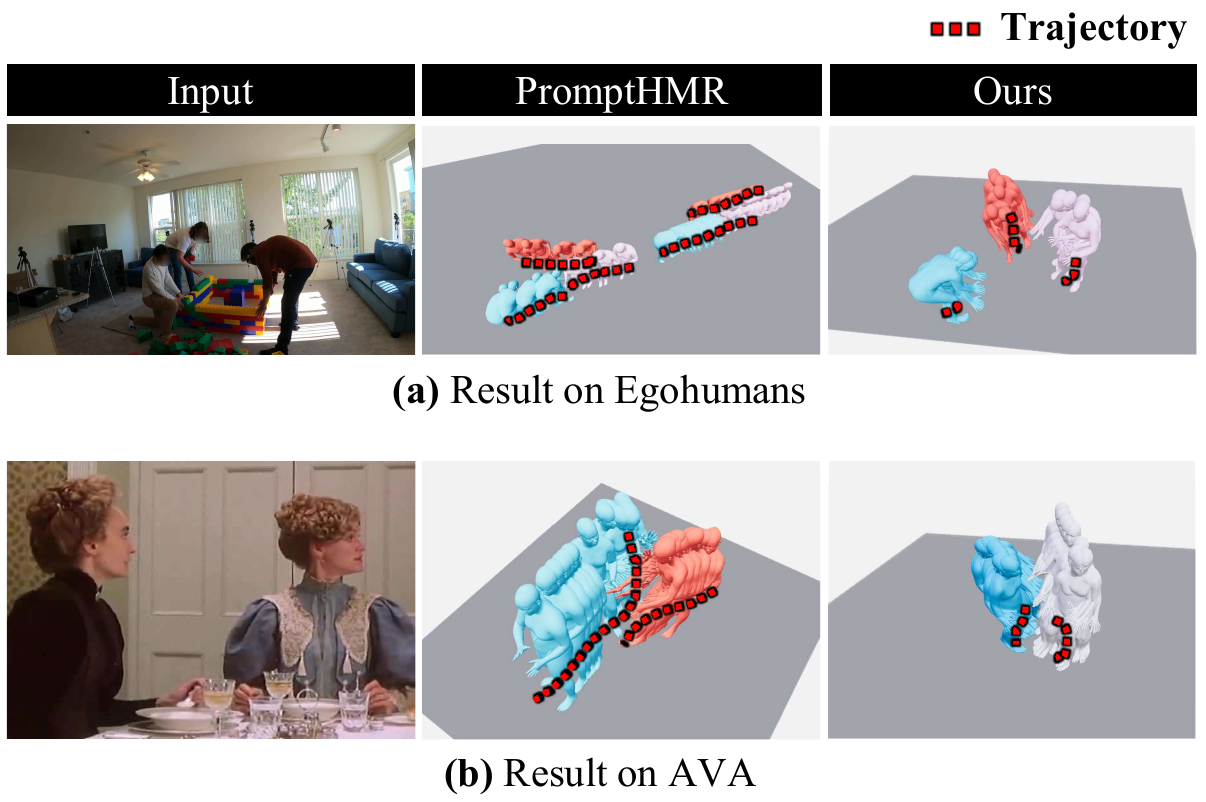}
    \caption{Qualitative comparison of global trajectories focusing on foot sliding artifacts. Compared to PromptHMR, our method produces smoother and more physically plausible motion with reduced foot sliding, as shown by the aligned trajectories (red dotted lines).}
   \label{fig:fig5}
   \vspace{-0.7cm}
\end{wrapfigure}

\noindent \textbf{Ablation study.} {%
Table~\ref{table:table3} presents an ablation study of each component of our method on EgoHumans~\cite{EgoHumans} using pose accuracy (W-MPJPE, MPJPE, MPVPE), temporal consistency (Accel), and camera accuracy (ATE). Removing the \textit{cam} module significantly increases both ATE and pose errors, emphasizing the importance of camera refinement. Excluding \textit{post} degrades temporal smoothness (Accel), validating its role in trajectory refinement. Disabling \textit{stage} causes performance drops across all metrics, underscoring the necessity of progressive optimization. The absence of \textit{align} causes the most significant performance decline, highlighting the critical importance of accurate human-scene alignment for cross-shot reconstruction. The full model (\textbf{Ours}) achieves the best performance across all metrics, confirming that each component contributes to the overall framework.
}
\section{Conclusion} {
In this paper, we propose a novel pipeline that performs human mesh reconstruction for multi-shot videos involving multiple-persons. A key contribution of our method is its ability to maintain identity consistency across shots successfully reconstructing meshes without losing person-specific identity information and achieving reliable re-identification. This is achieved by explicitly involving 3D scene reconstruction method and aligning 3D meshes to the shared 3D space. Experimental results demonstrated the effectiveness of our method and ablation studies confirmed the usefulness of each component.

\noindent \textbf{Limitation.} The primary objective of \abbrev~is to reconstruct continuous 3D trajectories, which inherently requires shot transitions to occur within the same physical environment (intra-scene). When a video undergoes an inter-scene transition with no spatial overlap, the concept of a continuous 3D motion becomes physically meaningless. While our geometry-driven Re-ID effectively handles intra-scene transitions, exploring identity tracking across entirely disjoint spaces remains an interesting future direction.

}

\section*{Acknowledgements}{
This work is supported by NRF grants (No. RS-2025-00521013 20\%, No. RS-2025-02216916 10\%) and IITP grants (No. RS-2020-II201336 Artificial intelligence graduate school program(UNIST) 10\%; No. RS-2025-25442824 AI Star Fellowship Program(UNIST) 10\%), funded by the Korean government (MSIT). This work is supported by CJ Corporation 50\%.
}

\bibliographystyle{splncs04}
\bibliography{main}

@String(CVPR= {IEEE Conf. Comput. Vis. Pattern Recog.})

@String(ICCV= {Int. Conf. Comput. Vis.})

@String(ECCV= {Eur. Conf. Comput. Vis.})

@String(TOG= {ACM Trans. Graph.})

@String(TIP  = {IEEE Trans. Image Process.})

@String(ICLR = {Int. Conf. Learn. Represent.})

@String(CVPR  = {CVPR})

@String(ICCV  = {ICCV})

@String(ECCV  = {ECCV})

@String(TOG   = {ACM TOG})

@String(TIP   = {IEEE TIP})

@String(ICLR  = {ICLR})

@inproceedings{paszke2019pytorch,
  title={{Pytorch: An imperative style, high-performance deep learning library}},
  author={Paszke, Adam and Gross, Sam and Massa, Francisco and Lerer, Adam and Bradbury, James and Chanan, Gregory and Killeen, Trevor and Lin, Zeming and Gimelshein, Natalia and Antiga, Luca and others},
  booktitle={NeuRIPS},
  year={2019}
}

@inproceedings{loshchilov2017decoupled,
  title={{Decoupled Weight Decay Regularization}},
  author={Loshchilov, Ilya and Hutter, Frank},
  booktitle={ICLR},
  year={2019}
}

@misc{distance,
  title = {Exact Euclidean distance transform},
  author = {{SciPy}},
  note = {Available at: \url{https://docs.scipy.org/doc/scipy/reference/generated/scipy.ndimage.distance_transform_edt.html}}
}

@inproceedings{rojas2025hamst3r,
  title={{Hamst3r: Human-aware multi-view stereo 3d reconstruction}},
  author={Rojas, Sara and Armando, Matthieu and Ghanem, Bernard and Weinzaepfel, Philippe and Leroy, Vincent and Rogez, Gregory},
  booktitle={ICCV},
  year={2025}
}

@inproceedings{choi2025hcp,
  title={{Humans as a calibration pattern: Dynamic 3d scene reconstruction from unsynchronized and uncalibrated videos}},
  author={Choi, Changwoon and Kim, Jeongjun and Cha, Geonho and Kim, Minkwan and Wee, Dongyoon and Kim, Young Min},
  booktitle={ICCV},
  year={2025}
}

@inproceedings{yang2025hac,
  title={{Humans as Checkerboards: Calibrating Camera Motion Scale for World-Coordinate Human Mesh Recovery}},
  author={Yang, Fengyuan and Gu, Kerui and Nguyen, Ha Linh and Tse, Tze Ho Elden and Yao, Angela},
  booktitle={ICCV},
  year={2025}
}

@inproceedings{yuan2025poses,
  title={{From poses to identity: Training-free person re-identification via feature centralization}},
  author={Yuan, Chao and Zhang, Guiwei and Ma, Changxiao and Zhang, Tianyi and Niu, Guanglin},
  booktitle={CVPR},
  year={2025}
}

@inproceedings{somers2024keypoint,
  title={{Keypoint promptable re-identification}},
  author={Somers, Vladimir and Alahi, Alexandre and Vleeschouwer, Christophe De},
  booktitle={ECCV},
  year={2024},
}

@inproceedings{tang2024mv,
  title={{MV-DUSt3R+: Single-Stage Scene Reconstruction from Sparse Views In 2 Seconds}},
  author={Tang, Zhenggang and Fan, Yuchen and Wang, Dilin and Xu, Hongyu and Ranjan, Rakesh and Schwing, Alexander and Yan, Zhicheng},
  booktitle={CVPR},
  year={2025}
}

@inproceedings{leroy2024mast3r,
    author = {Leroy, Vincent and Cabon, Yohann and Revaud, Jerome},
    title = {{Grounding Image Matching in 3D with MASt3R}},
    booktitle = {ECCV},
    year = {2024}
}

@article{ren2024grounded,
  title={Grounded sam: Assembling open-world models for diverse visual tasks},
  author={Ren, Tianhe and Liu, Shilong and Zeng, Ailing and Lin, Jing and Li, Kunchang and Cao, He and Chen, Jiayu and Huang, Xinyu and Chen, Yukang and Yan, Feng and others},
  journal={arXiv preprint arXiv:2401.14159},
  year={2024}
}

@inproceedings{wang2025prompthmr,
  title={{PromptHMR: Promptable Human Mesh Recovery}},
  author={Wang, Yufu and Sun, Yu and Patel, Priyanka and Daniilidis, Kostas and Black, Michael J and Kocabas, Muhammed},
  booktitle={CVPR},
  year={2025}
}

@inproceedings{rajasegaran2022tracking,
  title={Tracking People by Predicting 3{D} Appearance, Location \& Pose},
  author={Rajasegaran, Jathushan and Pavlakos, Georgios and Kanazawa, Angjoo and Malik, Jitendra},
  booktitle={CVPR},
  year={2022}
}

@inproceedings{goel2023humans,
    title={{Humans in 4{D}: Reconstructing and Tracking Humans with Transformers}},
    author={Goel, Shubham and Pavlakos, Georgios and Rajasegaran, Jathushan and Kanazawa, Angjoo and Malik, Jitendra},
    booktitle={ICCV},
    year={2023}
}

@inproceedings{ye2023slahmr,
    title={{Decoupling Human and Camera Motion from Videos in the Wild}},
    author={Ye, Vickie and Pavlakos, Georgios and Malik, Jitendra and Kanazawa, Angjoo},
    booktitle={CVPR},
    year={2023}
}

@InProceedings{hmrKanazawa17,
  title={{End-to-end Recovery of Human Shape and Pose}},
  author = {Angjoo Kanazawa and Michael J. Black and David W. Jacobs and Jitendra Malik},
  booktitle={CVPR},
  year={2018}
}

@InProceedings{shin2023wham,  
title={{WHAM: Reconstructing World-grounded Humans with Accurate 3D Motion}},
author={Shin, Soyong and Kim, Juyong and Halilaj, Eni and Black, Michael J.},  
booktitle={CVPR},  
year={2024}  
}

@inproceedings{farenzena2010person,
  author    = {Matteo Farenzena and Livio Bazzani and Alessandro Perina and Vittorio Murino and Marco Cristani},
  title     = {{Person Re-Identification by Symmetry-Driven Accumulation of Local Features}},
  booktitle = {CVPR},
  year      = {2010}
}

@inproceedings{gheissari2006person,
  author    = {Nouzhan Gheissari and Thomas B. Sebastian and Richard Hartley},
  title     = {{Person Reidentification Using Spatiotemporal Appearance}},
  booktitle = {CVPR},
  year      = {2006}
}

@inproceedings{ahmed2015improved,
  author    = {Ejaz Ahmed and Michael Jones and Tim K. Marks},
  title     = {{An Improved Deep Learning Architecture for Person Re-Identification}},
  booktitle = {CVPR},
  year      = {2015}
}

@inproceedings{varior2016gated,
  author    = {Rahul Rama Varior and Bing Shuai and Jiwen Lu and Dong Xu and Gang Wang},
  title     = {{A Gated Siamese Convolutional Neural Network Architecture for Human Re-Identification}},
  booktitle = {ECCV},
  year      = {2016}
}

@inproceedings{liu2017end,
  author    = {Huan Liu and Yuanhao Yu and Zhichao Zhou and Min Shao and Yun Fu},
  title     = {{End-to-End Comparative Attention Network for Person Re-Identification}},
  booktitle   = {TIP},
  year      = {2017}
}

@InProceedings{luo2019bag,
author = {Luo, Hao and Gu, Youzhi and Liao, Xingyu and Lai, Shenqi and Jiang, Wei},
title = {Bag of Tricks and a Strong Baseline for Deep Person Re-Identification},
booktitle = {CVPR},
year = {2019}
}

@article{dosovitskiy2020image,
  title={An image is worth 16x16 words: Transformers for image recognition at scale},
  author={Dosovitskiy, Alexey},
  journal={arXiv preprint arXiv:2010.11929},
  year={2020}
}

@inproceedings{he2021transreid,
  author    = {Shuting He and Hong-Xing Yu and Xiaoyu Shi and Xiao Zhang and Wei-Shi Zheng and Jian Sun},
  title     = {{TransReID: Transformer-Based Object Re-Identification}},
  booktitle = {ICCV},
  year      = {2021}
}

@inproceedings{kocabas2019vibe,
  author    = {Muhammed Kocabas and Nikos Athanasiou and Michael J. Black},
  title     = {{VIBE: Video Inference for Body Pose and Shape Estimation}},
  booktitle = {CVPR},
  year      = {2020}
}

@inproceedings{pavlakos2022human,
  title={Human mesh recovery from multiple shots},
  author={Pavlakos, Georgios and Malik, Jitendra and Kanazawa, Angjoo},
  booktitle={CVPR},
  year={2022}
}

@inproceedings{mehta2018single,
  title={Single-shot multi-person 3d pose estimation from monocular rgb},
  author={Mehta, Dushyant and Sotnychenko, Oleksandr and Mueller, Franziska and Xu, Weipeng and Sridhar, Srinath and Pons-Moll, Gerard and Theobalt, Christian},
  booktitle={3DV},
  year={2018},
}

@Inproceedings{pavlakos2022sitcoms3d,
  Title          = {The One Where They Reconstructed 3D Humans and Environments in TV Shows},
  Author         = {Pavlakos, Georgios and Weber, Ethan and Tancik, Matthew and Kanazawa, Angjoo},
  Booktitle      = {ECCV},
  Year           = {2022}
}

@inproceedings{AVA,
  author    = {Chunhui Gu and Chen Sun and David A. Ross and Carl Vondrick and Caroline Pantofaru and Yeqing Li and Sudheendra Vijayanarasimhan and George Toderici and Susanna Ricco and Rahul Sukthankar and Cordelia Schmid and Jitendra Malik},
  title     = {{AVA: A Video Dataset of Spatio-Temporally Localized Atomic Visual Actions}},
  booktitle = {CVPR},
  year      = {2018}
}

@inproceedings{Harmony4D,
  author    = {Khirodkar, Rawal and Song, Jyun-Ting and Cao, Jinkun and Luo, Zhengyi and Kitani, Kris},
  title     = {{Harmony4D: A Video Dataset for In-The-Wild Close Human Interactions}},
  booktitle = {NeurIPS},
  year      = {2024}
}

@inproceedings{EgoHumans,
  author    = {Khirodkar, Rawal and Bansal, Aayush and Ma, Lingni and Newcombe, Richard and Vo, Minh and Kitani, Kris},
  title     = {{EgoHumans: An Egocentric 3D Multi-Human Benchmark}},
  booktitle = {ICCV},
  year      = {2023}
}

@inproceedings{zhang2022egobody,
  title={Egobody: Human body shape and motion of interacting people from head-mounted devices},
  author={Zhang, Siwei and Ma, Qianli and Zhang, Yan and Qian, Zhiyin and Kwon, Taein and Pollefeys, Marc and Bogo, Federica and Tang, Siyu},
  booktitle={ECCV},
  year={2022},
}

@inproceedings{pavlakos2022multishot,
  author    = {Georgios Pavlakos and Angjoo Kanazawa and Jitendra Malik and Michael J. Black and Kostas Daniilidis},
  title     = {{Multi-Shot Person Tracking and Re-Identification in TV Series}},
  booktitle = {CVPR},
  year      = {2022}
}

@inproceedings{teed2021droid,
  title={{DROID-SLAM: Deep Visual SLAM for Monocular, Stereo, and RGB-D Cameras}},
  author={Teed, Zachary and Deng, Jia},
  booktitle={NeurIPS},
  year={2021}
}

@inproceedings{Mehta_2017,
   title={{VNect: real-time 3D human pose estimation with a single RGB camera}},
   booktitle={ACM ToG},
   author={Mehta, Dushyant and Sridhar, Srinath and Sotnychenko, Oleksandr and Rhodin, Helge and Shafiei, Mohammad and Seidel, Hans-Peter and Xu, Weipeng and Casas, Dan and Theobalt, Christian},
   year={2017}
}

@misc{pyscenedetect,
  author    = {Brandon, Castellano},
  title     = {{PySceneDetect: Automatic Scene Cut Detection Tool}},
  year      = {2014},
}

@inproceedings{SMPL:2015,
  author = {Loper, Matthew and Mahmood, Naureen and Romero, Javier and Pons-Moll, Gerard and Black, Michael J.},
  title = {{SMPL: A Skinned Multi-Person Linear Model}},
  booktitle = {ACM ToG},
  year = {2015}
}

@inproceedings{mueller2024hsfm,
  title={Reconstructing people, places, and cameras},
  author={M{\"u}ller, Lea and Choi, Hongsuk and Zhang, Anthony and Yi, Brent and Malik, Jitendra and Kanazawa, Angjoo},
  booktitle={CVPR},
  year={2025}
}

@article{Kuhn1955Hungarian,
  author = {Kuhn, Harold W.},
  journal = {Naval Research Logistics Quarterly},
  title = {{The Hungarian Method for the Assignment Problem}},

  year = 1955
}

@inproceedings{kim2025showmak3rcompositionaltvreconstruction,
  title={ShowMak3r: Compositional TV Show Reconstruction},
  author={Kim, Sangmin and Do, Seunguk and Park, Jaesik},
  booktitle={CVPR},
  year={2025}
}

@inproceedings{
  xu2022vitpose,
  title={Vi{TP}ose: Simple Vision Transformer Baselines for Human Pose Estimation},
  author={Yufei Xu and Jing Zhang and Qiming Zhang and Dacheng Tao},
  booktitle={NeurIPS},
  year={2022},
}

@inproceedings{shen2024gvhmr,
  title={{World-Grounded Human Motion Recovery via Gravity-View Coordinates}},
  author={Shen, Zehong and Pi, Huaijin and Xia, Yan and Cen, Zhi and Peng, Sida and Hu, Zechen and Bao, Hujun and Hu, Ruizhen and Zhou, Xiaowei},
  booktitle={ACM ToG},
  year={2024}
}

@InProceedings{wang2024dust3r,
    author    = {Wang, Shuzhe and Leroy, Vincent and Cabon, Yohann and Chidlovskii, Boris and Revaud, Jerome},
    title     = {{DUSt3R: Geometric 3D Vision Made Easy}},
    booktitle = {CVPR},
    year      = {2024},
}

@inproceedings{zhang2025humanmmglobalhumanmotion,
      title={{HumanMM: Global Human Motion Recovery from Multi-shot Videos}}, 
      author={Yuhong Zhang and Guanlin Wu and Ling-Hao Chen and Zhuokai Zhao and Jing Lin and Xiaoke Jiang and Jiamin Wu and Zhuoheng Li and Hao Frank Yang and Haoqian Wang and Lei Zhang},
      booktitle = {CVPR},
      year={2025},
}

@InProceedings{wang2025vggt,
    author    = {Wang, Jianyuan and Chen, Minghao and Karaev, Nikita and Vedaldi, Andrea and Rupprecht, Christian and Novotny, David},
    title     = {{VGGT: Visual Geometry Grounded Transformer}},
    booktitle = {CVPR},
    year      = {2025},
}

@inproceedings{SMPL-X:2019,
  title = {{Expressive Body Capture: 3D Hands, Face, and Body from a Single Image}},
  author = {Pavlakos, Georgios and Choutas, Vasileios and Ghorbani, Nima and Bolkart, Timo and Osman, Ahmed A. A. and Tzionas, Dimitrios and Black, Michael J.},
  booktitle = {CVPR},
  year = {2019}
}

%

\end{document}